# Evaluating the Potential of Leading Large Language Models in Reasoning Biology Questions

Xinyu Gong[1, †], Jason Holmes[3, †], Yiwei Li[2], Zhengliang Liu[2], Qi Gan[1], Zihao Wu[2], Jianli Zhang[1], Yusong Zou[1], Yuxi Teng[1], Tian Jiang[1], Hongtu Zhu[4], Wei Liu[3, *], Tianming Liu[2, *], Yajun Yan[1, *]

1. School of Chemical, Materials and Biomedical Engineering, College of Engineering, The University of Georgia, Athens, GA 30602, United States
2. School of Computing, The University of Georgia, Athens, GA 30602, United States
3. Department of Radiation Oncology, Mayo Clinic, Phoenix, AZ 85054, United States
4. Department of Biostatistics, University of North Carolina at Chapel Hill, Chapel Hill, NC 27599, United States

[†] These authors contributed equally to the work.

* Corresponding authors:

Wei Liu

5881 E Mayo Blvd, Phoenix, Mayo Clinic, AZ 85054, United States

Email: liu.wei@mayo.edu

Phone: +1-480-515-6296

Tianming Liu

420 Boyd Research and Education Center, The University of Georgia, Athens, GA, 30602, United States

Email: tliu@uga.edu

Phone: +1-706-542-3478

Yajun Yan

302 East Campus Road, The University of Georgia, Athens, GA, 30602, United States

Email: yajunyan@uga.edu

Phone: +1-706-542-8293

**Abstract**

Recent advances in Large Language Models (LLMs) have presented new opportunities for integrating Artificial General Intelligence (AGI) into biological research and education. This study evaluated the capabilities of leading LLMs, including GPT-4, GPT-3.5, PaLM2, Claude2, and SenseNova, in answering conceptual biology questions. The models were tested on a 108-question multiple-choice exam covering biology topics in molecular biology, biological techniques, metabolic engineering, and synthetic biology. Among the models, GPT-4 achieved the highest average score of 90 and demonstrated the greatest consistency across trials with different prompts. The results indicated GPT-4's proficiency in logical reasoning and its potential to aid biology research through capabilities like data analysis, hypothesis generation, and knowledge integration. However, further development and validation are still required before the promise of LLMs in accelerating biological discovery can be realized.

## 1. Introduction

In recent years, the field of biology has undergone a remarkable explosion of information and knowledge. Advanced sequencing technologies have efficiently generated vast datasets to underpin significant scientific discoveries [1, 2]. Meanwhile, CRISPR technology has revolutionized our capacity to precisely edit DNA in diverse organisms, greatly expediting our comprehension of gene functions and mutations [3, 4]. As cryo-electron microscopy continues to facilitate the elucidation of protein function and mechanisms, AI-assisted protein prediction is beginning to make noteworthy contributions [5, 6]. Collectively, these breakthroughs have deepened our understanding of life sciences and fueled the development of novel therapeutics, sustainable production methods, and more [7-11]. On the other hand, given the mounting data volume and the growing interdisciplinary complexity of knowledge, the accurate extraction, analysis, and integration of extensive biological information have become increasingly critical for fully harnessing the potential of these advancements and driving future innovations [12-14]. In this pursuit, adopting cutting-edge computational methods holds immense promise.

Recent advancements in artificial intelligence techniques hold promise for transforming biological research methods and uncovering new findings, given AI's documented success in addressing other complex scientific challenges. Large Language Models (LLMs) have emerged as one of the most promising approaches for solving biological problems. In some cases, LLMs could serve as powerful tools for generating protein sequences or as effective methods for acquiring 'content-aware' data representations from extensive sequence datasets using deep learning [15]. Applying these models enables gleaning insights from naturally occurring protein sequences encompassing known sequence diversity, while combining pre-existing structural and functional knowledge through multi-task learning [16, 17]. In addition, LLMs have been trained in the distribution of multiple biological molecules. By constructing training sets of more complex molecules, Flam-Shepherd et al. introduced a series of generative modeling tasks and demonstrated the power of LLMs in learning immense molecular distributions compared to most graph generative models [18]. Surprisingly, to overcome the real-world challenges of medical applications with LLMs, MultiMedQA, a benchmark comprising six medical question-answering datasets, was coupled with PaLM (Pathways Language Model) and evaluated as a helpful tool for clinical applications [19]. So far, LLMs have shown exciting applications in biology, allowing us to revolutionize various biological aspects by enabling efficient data analysis, hypothesis generation, drug discovery, and knowledge integration, ultimately accelerating advancements. Nevertheless, the potential of LLMs remains largely unexplored as continuous improvements in parameters, computing capabilities, datasets, and domain fine-tuning consistently expand their capabilities.

Biology can be interpreted through its own intrinsic language. In genomic sequences, scientists have observed and summarized certain patterns of DNA representing coding regions with certain functions [20]. In the language of genome, the gene family is the "word", and the coding region forms the "sentence". By analyzing and learning identified functional DNA sequences, algorithms can predict functions or families of input DNA sequences [21]. In the protein language, the grammar logic mirrors natural language. The amino acid sequence is the "letter", and the secondary structure forms the "word". The folded 3D protein structure is the "sentence", and the protein function convey the "meaning" [22]. Natural Language Processing (NLP) follows the grammar and logic of human language. It has potential to understand and speak biological language by embedding biological domains and fine-tuning LLMs. In this study, we designed 108 biological multiple-choice questions and tested the state-of-art LLMs, including GPT-3.5, GPT-4, PaLM2, Claude2, and SenseNova, for their capabilities in analyzing, solving, and answering biological questions by taking the exam. Among all tested LLMs, GPT-4 outperformed other LLMs by scoring an average score of 90 on the exam. The results indicated the supreme capability of GPT-4 in reasoning, analyzing, and answering biological questions, paving the way for LLMs to benefit biology research and education.

## 2. Related Work

### 2.1 Large Language Models

Natural Language Processing (NLP) integrates the subjects of linguistics and computer science to enable machines to learn, understand, and generate human language, serving as a crucial approach to achieving communication between humans and machines. To develop sophisticated language models, these models are trained on diverse linguistic datasets like literature, fiction, poetry, and research papers to analyze the linguistic structure and learn natural language usage [23]. In recent decades, advancements in computing power, high-quality datasets, and deep learning have enabled successful NLP applications like machine translation, sentiment analysis, information retrieval, and question-answering. As ChatGPT was unveiled in 2020, it caused a sensation across fields and boosted the development and launch of several LLMs [24]. Tech companies and scientific researchers now aim to create increasingly capable language models to incorporate into commercial products or scientific studies. Artificial General Intelligence (AGI) makes it straightforward to connect and implement these models [25].

Generative Pre-trained Transformer 3.5 (GPT-3.5) was developed by OpenAI and was first released in the chatbot, ChatGPT, enabling Artificial Intelligent Content Generation (AIGC) and human-like interaction. ChatGPT can perform different types of tasks, deliver fast responses, and provide comprehensive answers [26]. However, accuracy remains a concern. Recently, GPT-4 has emerged as the most advanced model of OpenAI. GPT-4 supersedes ChatGPT as a more reliable and creative version in reasoning capabilities. For example, GPT-4 achieved higher approximate percentiles (99th) in the Biology Olympiad than ChatGPT (31st) [27]. GPT-4 continues to improve through human feedback and real-world use. Alongside GPT, other competitive models are rushing out. Pathway Language Model 2 (PaLM2) is Google's latest multilingual model, which is highly capable of logical reasoning, coding, and mathematics while supporting over 100 different languages. PaLM API builds on PaLM2 and is fast at responding [28]. Claude 2 was developed by Anthropic as a rival chatbot to ChatGPT, mainly focusing on textual tasks with large context sizes of up to 100k tokens. It can work over hundreds of pages of documents or even books [29]. Recently, SenseNova was released as an emerging model capable of NLP, content generation, automated data annotation, and custom model training. SenseNova is the foundation model set of SenseTime with billions of parameters, working as the building block of their "foundation models + large-scale computing" systems [30].

## 2.2 Prompt Engineering

NLP models have occupied places in a range of domains such as medical consultation and code generation due to their powerful capability of processing natural languages [31, 32]. However, answering professional questions or performing research tasks requires extensive data and specific instructions to complete complex assignments. Recent studies show that employing proper prompts can efficiently yield high-quality results while saving computing time and resources [33]. Prompts are instructions that guide LLMs by providing customized rules and output formats. Well-designed prompts can enhance LLMs' capabilities and even enable solving sophisticated problems in specialized fields [34]. Thus, prompt engineering is critical for facilitating interaction between users and LLMs.

Since the quality of outputs is highly dependent on the quality of prompts, there is a growing demand for prompt engineering principles and skills. Besides, it remains challenging to efficiently engineer and fine-tune appropriate prompts in highly specialized domains. Prompts can be generated manually or automatically. Given that many complicated tasks require a huge number of prompts as guidance, adopting prompt functions to automatically generate prompt set and templates can improve working efficiency [35]. Several studies have addressed the significance of prompt engineering and proposed a series of catalogs and templates [33, 35, 36]. In this study, we presented a set of prompts and a chain of thoughts based on our examination in biology.

## 2.3 The Interdisciplinary of LLMs

The intersection of LLMs and various scientific domains, such as biological, medication, and ecology, etc., has sparked an interdisciplinary fusion with promising potential. In the current landscape, numerous surveys and initiatives are underway to explore the synergies between LLMs and other fields [19, 37, 38]. The integration of LLMs with biology is unraveling intricate insights within genomics, transcriptomics, and epigenomics [39]. By deciphering complex biological datasets and molecular interactions, LLMs' applications are advancing bioinformatics and enhancing our understanding of fundamental biological processes [40]. Researchers are diligently fine-tuning LLMs for biological contents, ensuring accurate comprehension of domain-specific terms and context. In addition to reading biology contexts, well-trained LLMs can also generate innovative small molecule distributions, nucleic acid sequences, and proteins with desired structures or functions. The surge in collaborative initiatives between computer scientists and biologists is cultivating a symbiotic learning process. The synergy, with the expanding reservoir of biological knowledge and the continuously evolving capabilities of LLMs, anticipates a future where interdisciplinary achievements are poised to drive revolutionary scientific progress.

In the field of medication, LLMs are aiding drug discovery by swiftly analyzing vast volumes of biomedical literature to explore and uncover hidden relationships between genes, proteins, and diseases. The combination of language processing with biological insights is also enabling personalized medicine, with the analysis of LLMs for individual patient data to predict optimal treatments. Med-PaLM performs encouragingly on the axes of our pilot human evaluation framework [19]. LLMs have emerged as a driving force in medical AI, holding significant promise to enhance the efficiency of clinical, educational, and research activities [38]. Several prominent chatbots, such as LaMDA (Google), GPT-3.5 (OpenAI) along with GPT-4 (OpenAI), have been extensively explored for medical applications [41]. For instance, GPT-4's capabilities have been investigated in generating medical notes from physician-patient transcripts have provided insights into innate medical knowledge, facilitating medical consultations, aiding in diagnoses, and contributing to medical education [41]. Beyond medicine, the interdisciplinary influence of LLMs extends to ecology and environmental studies by enabling data analysis to decipher intricate ecosystems and predict environmental changes [42]. Although the results are promising, these scientific domains are complex and growing. As LLMs maintain their significant involvement in research and everyday applications, assessing their performance gains heightened importance, particularly focusing on aspects such as safety, fairness, and bias [43].

## 3. Methods

### 3.1 Question Design

We developed a bank of 108 multiple-choice questions to assess conceptual understanding and reasoning abilities of LLMs in biology. The questions spanned key topics including:

- Fundamental molecular biology (59 questions): Key concepts such as DNA replication, transcription, translation, gene expression regulation, mutagenesis, etc.

- Common molecular biology techniques (17 questions): Principles and applications of biological techniques like PCR, DNA sequencing, DNA cloning, blotting methods, etc.

- Metabolic engineering (15 questions): Engineering microbial metabolism for bioproduction, metabolic flux analysis, optimizing pathway expression, etc.

- Synthetic biology (17 questions): Principles of genetic circuit design, toggle switches, oscillators, logic gates, CRISPR-based genome editing, etc.

The questions were designed to be intellectually challenging without relying on obscure facts. Questions went through expert review by two biology professors to confirm accuracy, relevance, and appropriate difficulty level.

### 3.2 Models Evaluation

We evaluated the performance of the following LLMs on the biology questions: GPT-4, GPT-3.5, PaLM2, Claude2, and SenseNova. These models represent the current state-of-the-art language model capabilities.

- GPT-4 [27] (OpenAI): One of the largest most capable generative pretrained multimodal models built on the transformer architecture, released by OpenAI on March 23, 2023. GPT-4 surpasses its predecessor GPT-3.5 in size and capability, delivering more coherent, contextually relevant, and nuanced outputs. GPT-4 has demonstrated human-level performance across several academic benchmarks, even achieving a score in the top 10% on a simulated bar exam. Notably, GPT-4 can process both text and images, extracting complex content and relationships from visual input.

- GPT-3.5 (OpenAI): Also known as ChatGPT, GPT-3.5 was released by OpenAI in November 2022. GPT-3.5 is a precursor of GPT-4 without image understanding abilities. Focused on and fine-tuned for conversational data, GPT-3.5 excels at generating dialogue responses. Its rapid global adoption garnered significant attention, not only from the NLP community but also across various scenarios, and it's continuously evolving for enhanced reasoning and safety.

- PaLM2 [44] (Google): Released by Google on May 23, 2023, PaLM2 is a general-purpose model that builds upon its predecessor, PaLM. Utilizing the Pathways model architecture, PaLM2 was trained on an expansive multilingual and diverse pre-training dataset encompassing human and programming languages, mathematical equations, scientific papers, and web content. It exhibits sophisticated reasoning capabilities across multiple languages and has spawned variants such as Med-PaLM 2 [45] and Sec-PaLM.

- Claude2 (Anthropic): Released by Anthropic on July 23, 2023, Claude2 is the successor to the Claude model, tailored for conversational applications. Its distinguishing feature is its extensive memory, capable of inputting up to 100k tokens, encompassing lengthy technical documentation or even entire books.

- SenseChat/SenseNova (SenseTime): Launched by Chinese AI company SenseTime on March 23, 2023, SenseChat builds on the SenseNova foundational model set. Trained extensively on a large-scale dataset with a particular emphasis on the Chinese context, it offers superior proficiency in processing texts. Additionally, SenseChat powers several innovative tools, including programming assistants, health consultation aids, and PDF reading assistants.

## 3.3 Testing Procedures

Each LLM was given the full set of 108 multiple choice biology questions. To account for variability due to prompt phrasing, each model was tested in 5 trials with differently worded prompts (Table 1), respectively. The prompts explained that the model should provide the single best answer for each question.

**Table 1 The prompts used in each trail**

| Trial | Prompts |
|---|---|
| Trial 1 | I am going to ask a multiple-choice question on a biological topic for the purpose of evaluating your understanding of the medical field. For the following question, provide the correct answer, one letter choice, without any justification. |
| Trial 2 | I want to evaluate your understanding of a biological topic by asking some multiple choice questions. Please answer the following question, designed for the purpose of evaluating your understanding of the topic. Provide the most correct letter choice (only one letter) without explanation. |
| Trial 3 | I'm a researcher that is trying to figure out how intelligent you are. Answer the following question as best you can. Return the best letter choice with no explanation. |
| Trial 4 | Please answer the following practice question as if you were a resident in training for a biological exam. Only give the correct answer in your response, a single letter choice. Do not explain your answer. |
| Trial 5 | We want to test your knowledge about the biological field. For this reason, we have created some questions to ask you. In your response, only report the correct answer (one letter) without explanation. |

### 3.4 Scoring and Analysis

In each trial, the number of questions answered correctly was recorded and scored for each LLM. Additionally, the consistency of each model's answers across the 5 trials was evaluated by comparing the specific questions answered correctly. Mean scores and variability across trials were calculated and graphed for each model. Comparative performance was assessed based on both aggregate scores and answer stability.

### 4. Results

#### 4.1 Comparison between different LLMs in performing biology multiple-choice exam

The development of LLMs has demonstrated a significant impact across diverse areas, including education [46], medication [47], and agriculture [48]. The breakthrough of applying LLMs in other fields depends on embedding and fine-tuning the specific domains into the models [49]. As the reasoning capability was improved, the state-of-the-art LLMs gradually showed enhanced ability in analyzing and solving different types of questions and were able to substitute humans for some simple tasks. The multiple-choice section is an important approach for validating the mathematical and logical reasoning skills of LLMs. To test the capabilities of the current competitive LLMs for biological reasoning, we designed over 100 biological multiple-choice questions covering fundamental molecular biology concepts, common molecular biology techniques, metabolic engineering, and synthetic biology. In this study, we tested five popular models, namely GPT-4, GPT-3.5, PaLM2, Claude2, and SenseNova, benchmarking their reasoning capabilities in the biology field. We conducted five trials with different-phrased prompts to mitigate prompt engineering effects on each model. In each trial, the performance of each model was shown in parallel, and the correct answers were marked in colored square (Figure 1a). The results indicated that all models could deal with most of the questions, but GPT-4 consistently outperformed other models by getting more colored markers. Remarkably, among all five trials, GPT-4 presented the most stable pattern of marked correct answers than other counterparts.

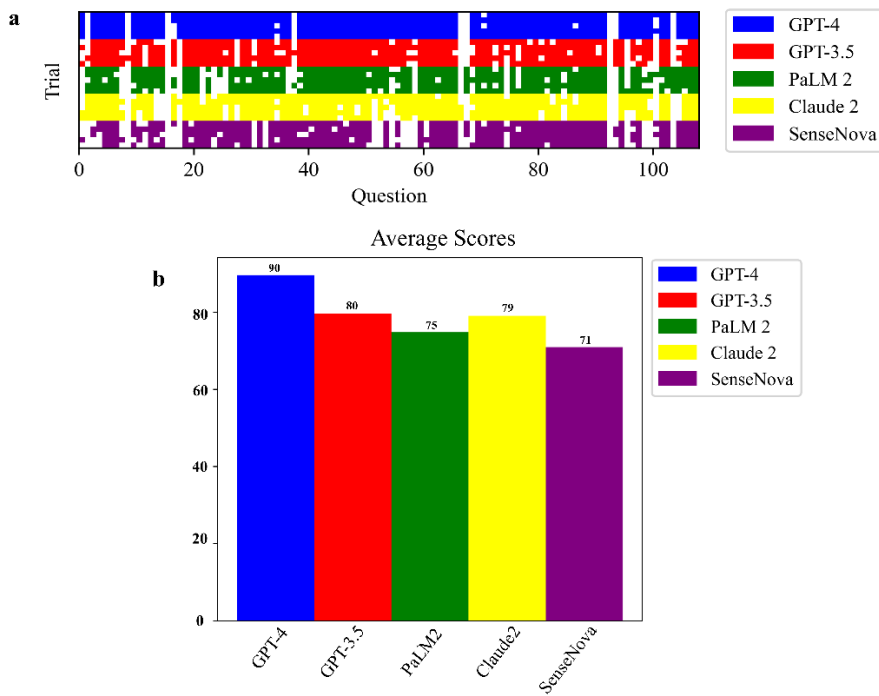

**Figure 1 Overall performance of five LLMs in the biological exam.** (a) Raw data for each LLM with five trails with different-phrased prompts, where the rows represent separate trails, and the columns represent the test questions. Color shaded squares indicate correct answers, while white

squares indicate the incorrect answers. (b) Average test scores for each LLM in the biological multiple-choice question exam across five trails.

Additionally, our initial analysis of the five LLMs across five trials revealed overall satisfactory performance, with all models scoring above 70 points on average (Figure 1b). However, GPT-4 clearly stood out with a significantly higher average score of 90 points compared to the 70–80-point range of the other models. To further analyze consistency, we calculated the standard deviation and inter-trial correlation for each model (Figure 2). GPT-3.5, PaLM2, Claude2, and SenseNova performed comparably, with standard deviation of 0.4-0.45 and correlation of 0.57-0.67. In contrast, GPT-4 demonstrated higher confidence and consistency, with a lower standard deviation of 0.30 and a higher correlation of 0.87. Next, since we categorized the questions into four subtypes, the average score and standard deviation were also calculated to evaluate the capabilities of LLMs in different types of questions (Figure 3). Overall, GPT-4 outperformed the other LLMs across all subcategories. Notably, GPT-4 was especially good at reasoning questions in metabolic engineering and synthetic biology, achieving 100 mean points and 92 mean points, respectively (Figure 3a). Meanwhile, GPT-4 presented high consistency by showing lower standard deviation than the counterparts under each subcategory (Figure 3b). These results indicated GPT-4's superior confidence and alignment in reasoning biology questions.

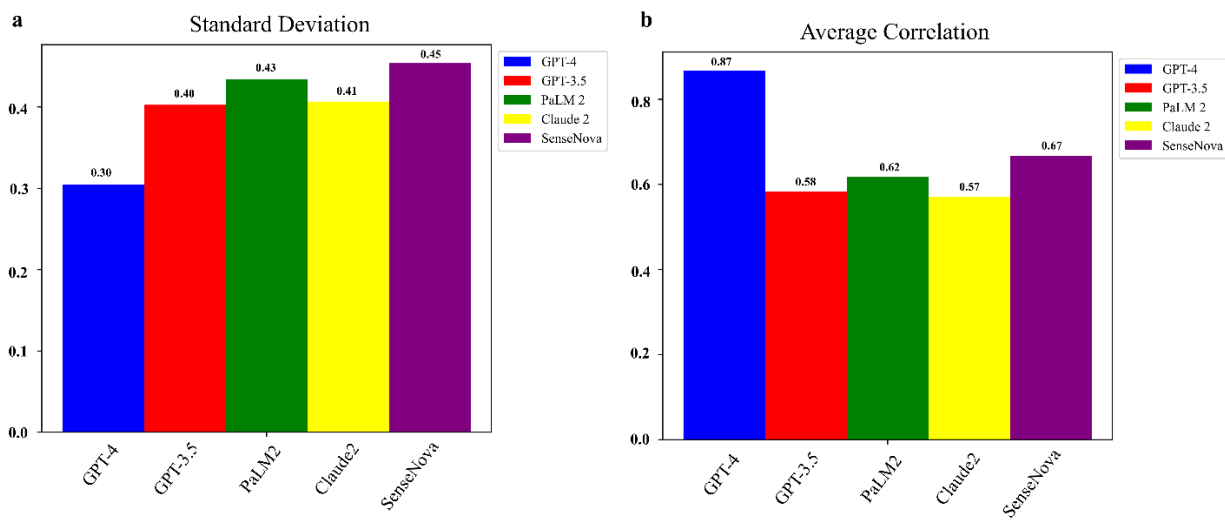

**Figure 2 Quantification of the overall consistency of scoring.** (a) Standard deviation of total scores for each LLM across five trails. (b) Average correlation between trials for each LLM.

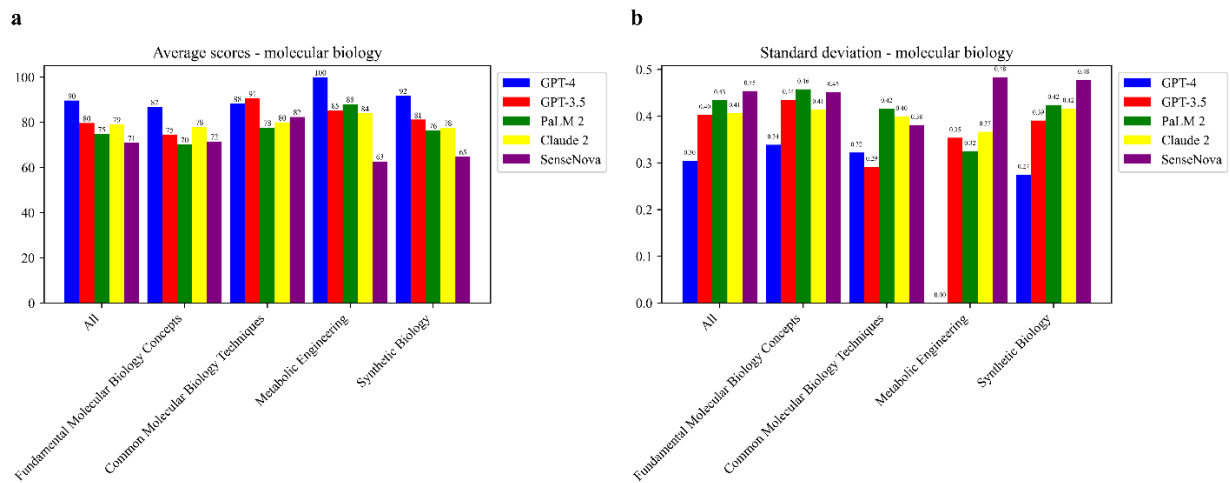

**Figure 3 The performance of five LLMs by category.** (a) Average scores for each LLM by category. (b) Standard deviation for each LLM by category.

To evaluate the degree of confidence in the answers given by the LLMs, we compared the results of correct answer occurrences per question for each model with the expected distribution that would occur if the models were guessing at random (Figure 4). All LLMs showed trends opposite to the random guessing. In comparison, PaLM2 and SenseNova were less confident, getting over 50% of questions correct in each trial (Figure 4c and Figure 4e). GPT 3.5 and Claude 2 were more confident, getting over 60% of questions correct in each trial (Figure 4b and Figure 4d). GPT-4 was undoubtedly the most confident model, getting over 80% of questions correct in each trial and less than 10% of questions incorrect (Figure 4a). Through this degree of confidence test, GPT-4 demonstrated proficient biological reasoning, indicating the potential in biology applications.

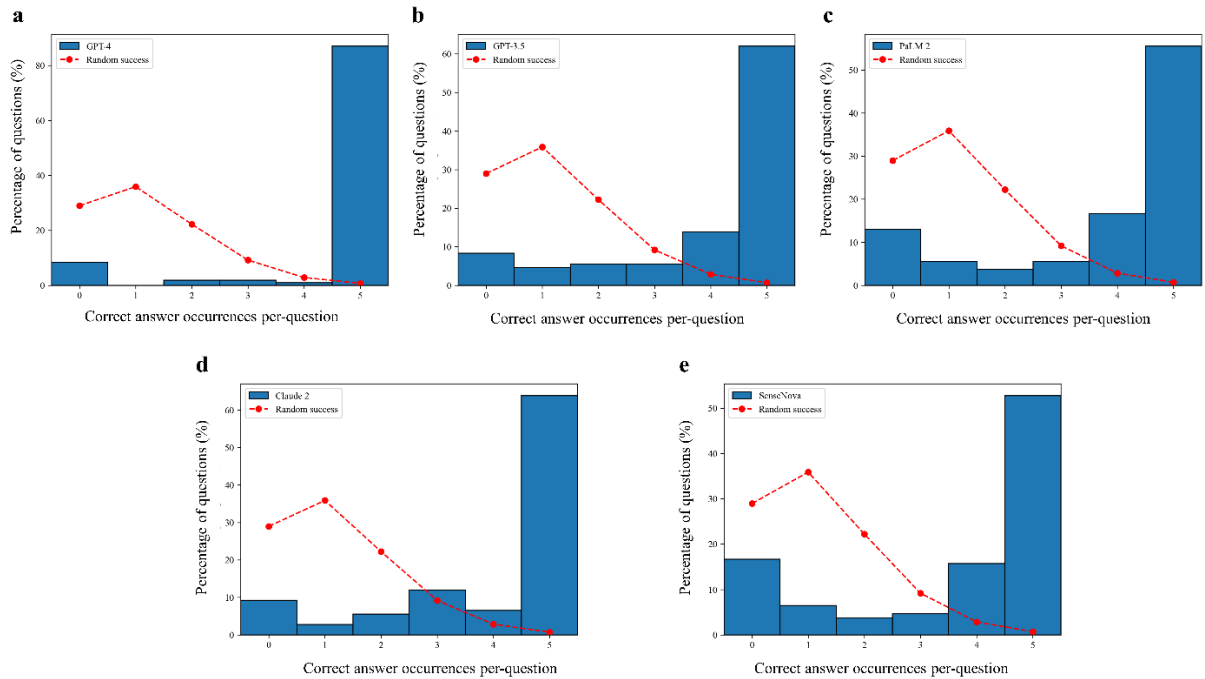

**Figure 4 Quantification the degree of confidence in the answers.** The blue bars in figures (a-e) describe the number of correct answer occurrences per-question for each LLM. The dashed red lines in figures (a-e) present the expected number of correct answers in 5 trials, when randomly guessing based on the Poisson distribution. (a) degree of confidence test of GPT-4. (b) degree of confidence test of GPT-3.5. (c) degree of confidence test of PaLM2. (d) degree of confidence test of Claude 2. (e) degree of confidence test of SenseNova.

## 5. Discussion

Faced with energy shortage, environmental problems, and healthcare issues, biological strategies are blooming solutions to achieve bio-fuel production, pharmaceutical synthesis, and disease diagnosis. Recent advances in computational methods and their integration with biology have accelerated research progress in fields such as genome mining, protein structure prediction, and protein engineering. We can foresee the bright future of AI-aided biology. LLMs present opportunities to enhance understanding and interpretation of biological language. Fine-tuning LLMs on domain-specific corpora enables applications in aiding rational biological experimental design and providing foundational biological knowledge to non-experts seeking training. However, as an emerging field, our current comprehension of biology using AI remains limited in depth and scope. Biological knowledge continuously expands at a rapid pace, making it difficult to immediately develop specialized AI systems that fully capture the nuances of the field. Therefore, continually training LLMs on high-quality biological datasets through a lifelong learning approach may maximize their utility for advancing scientific research and education.

In this study, evaluation of answer patterns revealed GPT-4 had the highest confidence in its responses, getting over 80% of questions correct per trial. In contrast, the other models showed lower confidence levels. GPT-4's high confidence indicates its suitability for deploying in applications requiring reliable outputs. On the other hand, testing prompt variations for each model showed GPT-4 had the most consistent performance, underscoring the importance of prompt engineering to optimize LLMs. Further research into prompt design strategies tailored for biology questions would be beneficial.

Currently, the language processing capabilities of LLMs could aid the analysis of large-scale biological datasets like genomics, transcriptomics, and proteomics. LLMs can help extract insights from massive amounts of sequencing data as well as integrate and interpret findings across datasets. Fine-tuning biology corpora can teach LLMs the language of biological data. The applications of LLMs in protein design leverage the understandings of sequence-function relationships and design principles. LLMs can propose solutions to engineering challenges and accelerate iterative design cycles. They may also uncover novel strategies human engineers may not conceive of.

Additionally, LLMs present opportunities to automate aspects of literature analysis which currently require extensive human effort. LLMs can rapidly review vast bodies of text to extract key findings, synthesize conclusions across papers, and even generate novel hypotheses. This could significantly accelerate knowledge discovery in biology. LLMs like chatbots can provide on-demand tutoring and practice for students learning biology concepts. They can also enable those without formal biology training to query biological knowledge interactively. Fine-tuned LLMs can become powerful educational resources to make biology knowledge more accessible. Advancing LLMs specialized in biology promises to catalyze discovery across diverse research areas and make biological insights more accessible. Overall, the creative knowledge integration abilities of

LLMs could enable breakthroughs in biological understanding and experimentation exceeding human limitations.

## 6. Conclusion

This study illuminates the emerging potential of LLMs to advance biology research and education through their language processing capabilities. The leading LLM, GPT-4, exhibited proficiency in logical reasoning applied to conceptual biology questions, outperforming other state-of-the-art models. While further validation is required, the results point to exciting possibilities for LLMs to accelerate knowledge discovery by analyzing massive biological datasets, streamlining literature analysis, improving molecular engineering, broadening access to biology education, and synergizing with human capabilities. With proper oversight and governance to ensure safety and ethics, harnessing LLMs to complement human intelligence could catalyze breakthroughs, leading to advancements in biomedicine, biotechnology, bioengineering, and sustainability. Further interdisciplinary research that brings together biologists, computer scientists, and experts across fields will be key to transforming these potentials into realities that benefit science and society.

**Author contributions**

XG, JH, TL, and YY contributed to conception and design of the study. XG, JZ, QG, YZ, YT and TJ designed the 108-question exam. JH, YL, LZ, WL, and TL conducted the experimental design and model running. JH and XG performed all data analysis. XG guided the writing process and wrote the initial draft. QG, JZ, YZ, and YT contributed the writing to the introduction and related work of the manuscript. ZL, ZW, HZ, and TL advised on LLMs concepts and contributed writing to the abstract method, discussion, and conclusion of the manuscript. All authors contributed to the article and approved the submitted version.

**Conflict of interest**

The authors declare no competing financial interest.

## Supporting Information

The supplementary file contains the original 108 multiple-choice questions on the biology topic.

## Acknowledgement

We acknowledge the support from the College of Engineering, The University of Georgia, Georgia, Athens, United States.